\pdfoutput=1
\documentclass[11pt]{article}

\usepackage[final]{acl}
\usepackage{xcolor}
\usepackage{graphicx}
\usepackage{colortbl}
\usepackage{adjustbox}
\usepackage{times}
\usepackage{booktabs}
\usepackage{multirow}
\usepackage{latexsym}
\usepackage{booktabs} % 提供 \toprule, \midrule, \bottomrule
\usepackage{colortbl} % 提供 \rowcolor 和单元格背景色
\usepackage{multirow} % 提供 \multirow
\usepackage{xcolor}  
\usepackage[T1]{fontenc}
\usepackage{rotating} % For the sideways environment
\usepackage{pifont}   % For checkmark symbol% This assumes your files are encoded as UTF8
\usepackage[utf8]{inputenc}
\usepackage{microtype}
\usepackage{mathtools} % 添加 mathtools 宏包

\usepackage{inconsolata}
\usepackage{tabularx}
\usepackage{graphicx}
\usepackage{subcaption} % 用于创建子表格
\usepackage{colortbl}
\usepackage{amsfonts}
\usepackage{amsmath}

\usepackage{authblk}
\usepackage[normalem]{ulem}

\title{HSCR: Hierarchical Self-Contrastive Rewarding for Aligning Medical Vision Language Models}
\author{
  \textbf{Songtao Jiang\textsuperscript{1}},
  \textbf{Yan Zhang\textsuperscript{2}},
  \textbf{Yeying Jin\textsuperscript{3}},
  \textbf{Zhihang Tang\textsuperscript{1}}
\\
  \textbf{Yangyang Wu\textsuperscript{1}},
  \textbf{Yang Feng\textsuperscript{4}},
  \textbf{Jian Wu\textsuperscript{1,5}},
  \textbf{Zuozhu Liu\textsuperscript{1,5†}}
\\
\\ % Double backslash for a larger gap before affiliations as in the example
  \textsuperscript{1}Zhejiang University,
  \textsuperscript{2}Byte Dance,
  \textsuperscript{3}National University of Singapore
\\
  \textsuperscript{4}Angelalign Inc., China
  \textsuperscript{5}Zhejiang Key Laboratory of Medical Imaging Artificial Intelligence
\\
  \small{
    \textbf{Correspondence\textsuperscript{†}:} \href{zuozhuliu@intl.zju.edu.cn}{zuozhuliu@intl.zju.edu.cn} 
  }
}

\begin{document}
  \maketitle
\begin{abstract}

Medical Vision-Language Models (Med-VLMs) have achieved success across various tasks, yet most existing methods overlook the modality misalignment issue that can lead to untrustworthy responses in clinical settings. 
In this paper,  we propose Hierarchical Self-Contrastive Rewarding (HSCR), a novel approach that addresses two critical challenges in Med-VLM alignment: 1) Cost-effective generation of high-quality preference data; 2) Capturing nuanced and context-aware preferences for improved alignment. 
HSCR first leverages the inherent capability of Med-VLMs to generate dispreferred responses with higher sampling probability. By analyzing output logit shifts after visual token dropout, we identify modality-coupled tokens that induce misalignment and derive an implicit alignment reward function. This function guides token replacement with hallucinated ones during decoding, producing high-quality dispreferred data. Furthermore, HSCR introduces a multi-level preference optimization strategy, which extends beyond traditional adjacent-level optimization by incorporating nuanced implicit preferences, leveraging relative quality in dispreferred data to capture subtle alignment cues for more precise and context-aware optimization. Extensive experiments across multiple medical tasks, including Med-VQA, medical image captioning and instruction following, demonstrate that HSCR not only enhances zero-shot performance but also significantly improves modality alignment and trustworthiness with just 2,000 training entries. Code is released on \href{https://github.com/jiangsongtao/HSCR}{https://github.com/jiangsongtao/HSCR}.

\end{abstract}
\section{Introduction}

Medical Vision-Language Models (Med-VLMs) have shown strong performance in tasks like medical visual question answering (Med-VQA)~\citep{moor2023med,saab2024capabilities,gai-etal-2025-medthink,li2024llava,jiang2025omniv,XU2025126585} by integrating pre-trained vision encoders into large language models (LLMs), allowing access to visual information. However, limited quality and quantity of paired multimodal medical training data often lead to modality misalignment~\citep{jiang2024med,krieger1992overcoming}. As a result, medical VLMs may {hallucinate image contents}, favoring text-based preferences over actual visual content. This misalignment undermines trustworthiness, posing challenges for reliable applications of these models in high-stakes medical scenarios~\citep{liu2023medical}.
\begin{figure}
    \centering
    \includegraphics[width=1\linewidth]{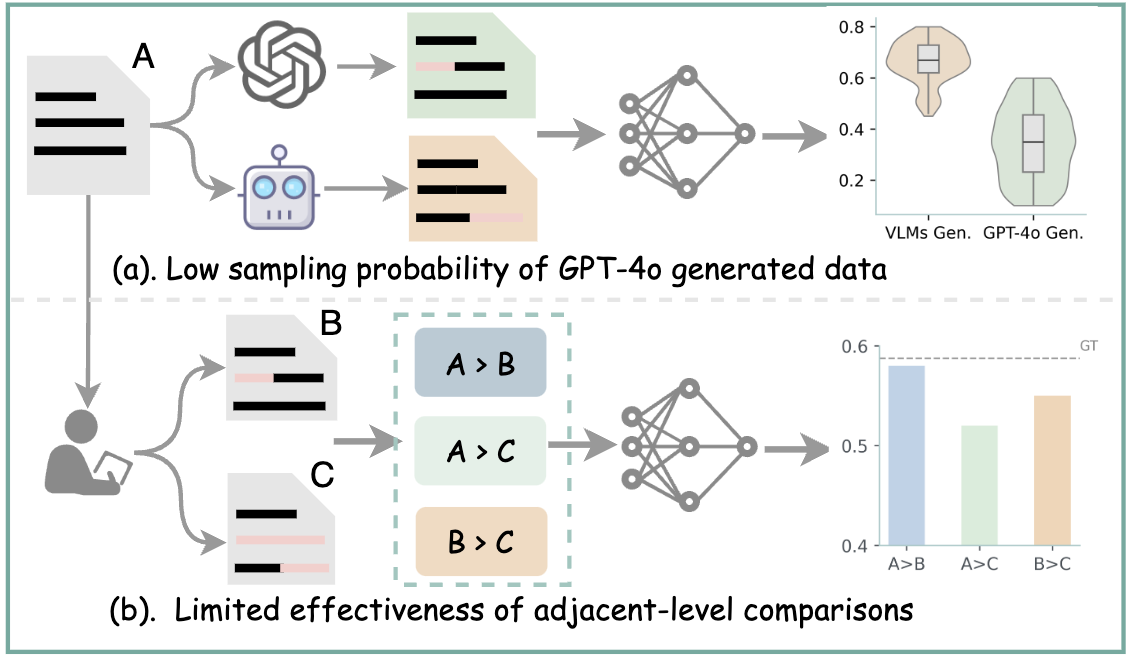}
\caption{Two key challenges in preference optimization for Med-VLMs. See Appendix~\ref{details:fig1} for details.}
    \label{fig:intro1}
\end{figure}

Recent work in multimodal learning has explored preference optimization methods to improve modality alignment, such as Reinforcement Learning with Human Feedback~\citep{sun2023aligning,song2024preference} and Direct Preference Optimization~\citep{rafailov2024direct,jiang2024modality,zhou2024aligning,zhang2024automated}. While these approaches show promise in general domains, their application to Med-VLMs faces two critical challenges (Figure~\ref{fig:intro1}): (1) limited sampling probabilities for preferred/dispreferred responses during optimization, and (2) reduced effectiveness of adjacent-level comparisons in weakly trained Med-VLMs. These issues stem from significant gaps in data quality, scale, and distribution between general VLMs and Med-VLMs, compounded by reliance on manual annotations or synthetic preference data from larger VLMs like GPT-4o~\citep{hurst2024gpt}.

For the first challenge, human-annotated or GPT-4o-generated preference data exhibit limited sampling probabilities during Med-VLM optimization. This inefficiency arises from a misalignment between the decoding behavior of Med-VLMs and the external preference data distribution (e.g., GPT-4o), primarily due to divergent training data sources and objectives between GPT-4o and Med-VLMs. Consequently, preferred responses are rarely sampled during optimization, resulting in weak reward signals and suboptimal alignment performance~\citep{zhou2024calibrated,azar2024general}.
For the second challenge, adjacent-level comparisons between correct and incorrect responses often exhibit pronounced disparities, causing Med-VLMs to easily saturate their ability to distinguish preferred and dispreferred outputs. This phenomenon limits their capacity to learn nuanced preferences, as large preference gaps also obscure clear optimization directions during training~\citep{zhou2024beyond}. The issue is exacerbated in medical domains, where subtle distinctions between plausible responses require finer-grained learning.

\begin{figure*}[t!]
    \centering
    \includegraphics[width=1\linewidth]{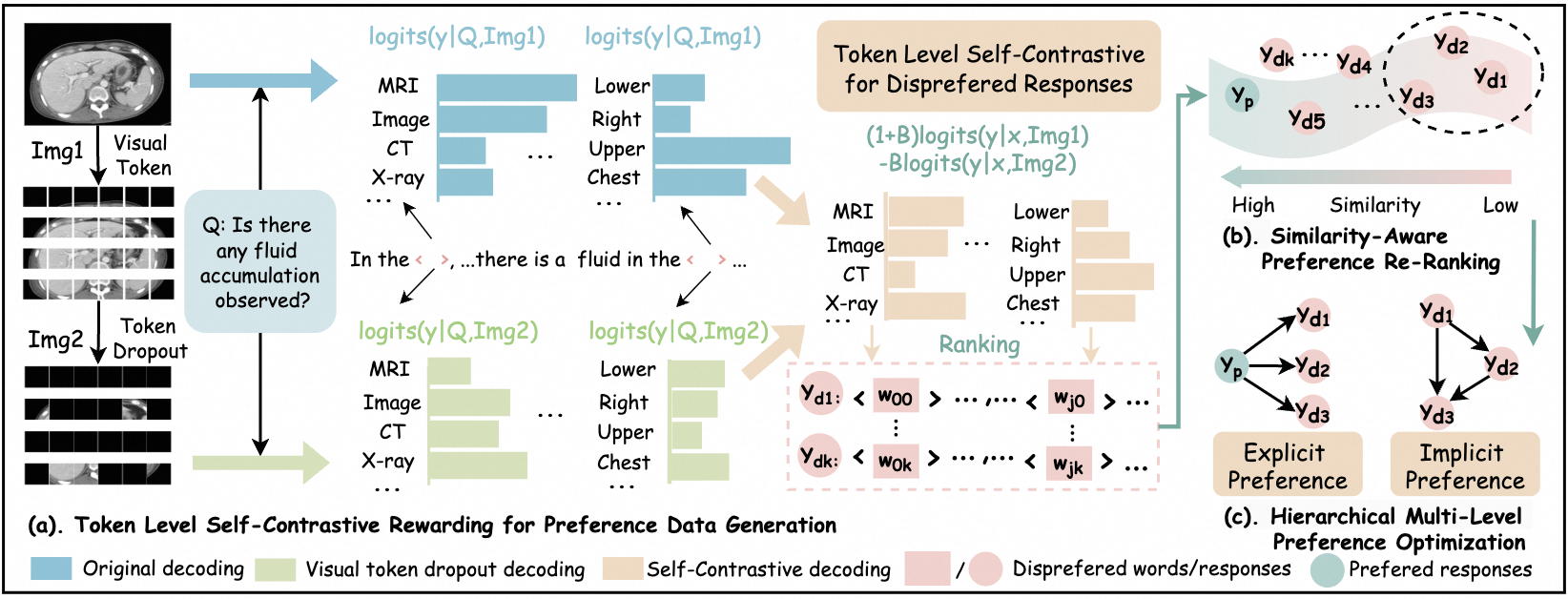}
\caption{Overview of the HSCR pipeline, comprising token-level self-contrastive rewarding for data generation (Methods~\ref{methods1}), similarity-aware preference re-ranking for quality control (Methods~\ref{methods2}), and hierarchical multi-level preference optimization (Methods~\ref{methods3}).}
    \label{fig:pipeline}
\end{figure*}

To address these challenges, we propose Hierarchical Self-Contrastive Rewarding (HSCR), a novel preference optimization method for Med-VLM alignment. HSCR encompasses three steps: token-level self-contrastive rewarding for data generation, similarity-aware preference re-ranking, and multi-level preference optimization. First, to generate preference data with enhanced sampling probabilities, we leverage the inherent capabilities of Med-VLMs to produce misaligned responses, eliminating the need for external resources or annotations. These misaligned responses naturally exhibit higher sampling probabilities, making them effective examples of dispreferred data. In particular, inspired by the masking strategies in Masked Autoencoders~\citep{he2022masked} and Vision Transformers~\citep{dosovitskiy2020image},we introduce visual token dropout to expose inherent misalignment in the model's generation behavior. By analyzing the resulting differences in logits, we can identify strongly modality-coupled tokens that are prone to inducing hallucinations due to significant shifts in their logits. These logit differences are formed as an implicit reward function to replace sensitive tokens with hallucinated ones. Afterwards, to ensure that response rankings accurately reflect semantic differences, we compute the semantic similarity between dispreferred and preferred responses, re-ranking them accordingly. This process yields rank-based preference lists.

Finally, unlike existing preference optimization techniques that focus solely on categorizing correct responses as preferred and hallucinated responses as dispreferred (i.e., explicit preference), our study reveals that varying levels of incorrectness in dispreferred responses can provide richer and more nuanced preference signals. Specifically, we designed an implicit optimization objective which encourages the model to discern differences in the degree of incorrectness among dispreferred responses. This multi-level joint preference optimization approach not only captures broad, high-level alignment signals through explicit preference but also learns subtle and intricate preferences through implicit preference.
HSCR achieves performance improvements across a wide range of Med-VQA, captioning, and instruction-following tasks. Notably, it achieves state-of-the-art (SOTA) zero-shot performance on Rad-VQA~\citep{lau2018dataset}, SLAKE~\citep{liu2021slake}, and PathVQA~\citep{he2020pathvqa}. Code and datasets will be released. 

\section{Preliminaries}
Direct Preference Optimization (DPO) streamlines alignment by directly optimizing the policy \( \pi_{\eta} \) using preference data \( \mathcal{P} \), eliminating the need for an explicit reward model~\citep{rafailov2024direct}. DPO links the reward function \( g(x, y) \) to the policy:
\begin{equation}
\label{DPOr}
    g(x, y) = \gamma \log \frac{\pi_{\eta}(y|x)}{\pi_{\text{base}}(y|x)} + \gamma \log Z(x),
\end{equation}
where \( Z(x) \) is the partition function. The optimization objective is:
\begin{equation}
\resizebox{0.5\textwidth}{!}{$
\begin{aligned}
   \mathcal{J}_{\text{DPO}}(\pi_{\eta}; \pi_{\text{init}}) &= - \mathbb{E}_{(x, y_w, y_l) \sim \mathcal{P}} \left[ \log \sigma \left( \gamma \log \frac{\pi_{\eta}(y_w|x)}{\pi_{\text{init}}(y_w|x)} \right. \right. \\
   &\quad \left. \left. - \gamma \log \frac{\pi_{\eta}(y_l|x)}{\pi_{\text{init}}(y_l|x)} \right) \right], 
\end{aligned}
$}
\end{equation}
where \( y_w \) and \( y_l \) denote preferred and less preferred outputs, respectively. This approach enhances efficiency and precision in alignment~\citep{dong2024rlhf,pal2024smaug}. Details are in Appendix~\ref{details:pre}.

\section{Hierarchical Self-Contrastive Rewarding (HSCR)}
In this section, we delve into the HSCR framework (see Fig.~\ref{fig:pipeline}). First, we construct preference datasets without external tools through self-contrastive rewarding. Next, we perform semantic similarity-based checking and re-ranking of the preference data. Finally, we optimize a supervised fine-tuned medical VLM, modelled as a policy $\pi_\theta$ parameterized by $\theta$, using multi-level preference optimization to achieve enhanced alignment.

\subsection{Token-Level Self-Contrastive Rewarding for Data Generation}
\label{methods1}
For data generation, we treat the ground truth as the preferred response $y_w$ and aim to construct dispreferred responses $y_l$ that reflect the inherent misalignment or unreliable behavior of VLMs. To achieve this, we adopt a two-step approach. First, to expose potential misalignment, we employ a visual token dropout strategy (see Fig.~\ref{fig:pipeline} a) to disrupt the image modality. Specifically, given the original visual token $i$, we apply a 70\% dropout rate to obtain $i'$. Combined with the given textual query $x$, we compute the token logits output by the VLM as $\text{logit}_\theta(y \mid i, x)$ and $\text{logit}_\theta(y \mid i', x)$, respectively. Second, to identify tokens prone to causing misalignment, we locate the top-$n$ tokens with the largest logit differences between the two distributions. These tokens, which exhibit strong modality coupling, are often error-prone:
\begin{equation}
\resizebox{0.5\textwidth}{!}{$
\begin{aligned}
P_{\text{diff}} = \text{Softmax} \left[ (1 + \beta) \cdot \text{logit}_\theta(y \mid i, x) \right. 
- \left. \beta \cdot \text{logit}_\theta(y \mid i', x) \right],
\end{aligned}
$}
\end{equation}
where $\beta$ controls the contrast strength between distributions, with higher values enhancing the distinction between the two. After identifying these sensitive tokens, we generate token-level dispreferred responses by replacing them with incorrect tokens through contrastive decoding. Specifically, for each sensitive token, we decode based on the logit differences from $P_{\text{diff}}$ in ascending order, substituting them with tokens that exhibit lower logit differences. These substituted tokens, which are weakly correlated with the actual visual information, often correspond to hallucinated outputs of the model~\citep{leng2024mitigating}. By replacing all sensitive tokens, we generate a set of dispreferred responses with varying degrees of incorrectness:
\begin{equation}
\{y_{l1}, y_{l2}, \dots, y_{lk}\} \sim  \left( \prod_{t=1}^{T} P_{\text{diff}}(y_t \mid y_{<t}, i, i', x) \right),
\end{equation}
where $y_t$ is the token at position $t$, $y_{<t}$ denotes preceding tokens, and $T$ is the sequence length (Hyperparameters are in Section~\ref{ref:expe_detail}).

\subsection{Similarity-Aware Preference Re-Ranking}
\label{methods2}
After applying the approach in Section~\ref{methods1}, we obtain a set of candidate dispreferred responses. However, these responses may exhibit varying degrees of semantic similarity to the preferred response. For instance, some dispreferred responses might be partially correct or contextually relevant but less precise, while others could be entirely unrelated or misleading. To ensure accurate preference ordering, we introduce a similarity-aware re-ranking module to refine the responses (see Fig.~\ref{fig:pipeline} b). Specifically, we compute the semantic similarity~\citep{corley2005measuring} between each dispreferred response $y_{lk}$ and the preferred response $y_w$, denoted as $\text{sim}(y_{lk}, y_w)$. The responses are then re-ranked in descending order of similarity. From the re-ranked list, we select \( j \) responses such that their similarity differences with respect to \( y_w \) are at least \( 0.1 \) for optimization. By explicitly integrating semantic similarity into the ranking process, this approach enables a more nuanced distinction between responses of varying quality. Consequently, the final dispreferred responses $\{y_{l1}, y_{l2}, \dots, y_{lj}\}$ will be used for subsequent preference optimization.

\subsection{Hierarchical Multi-Level Preference Optimization}
\label{methods3}
To comprehensively optimize both explicit and implicit preferences in training data, we propose {Multi-Level Preference Optimization (MLPO)}, a novel framework that enhances the model's ability to distinguish not only between preferred and dispreferred responses but also among dispreferred responses at varying levels. Unlike traditional binary preference optimization methods~\citep{yu2024rlhf}, which rely on coarse-grained comparisons, MLPO enables fine-grained preference learning through two key components: \textit{Explicit Preference Learning} and \textit{Implicit Preference Learning}, each performing multi-level optimization (see Fig.~\ref{fig:pipeline} c).

\noindent\textbf{Explicit Preference Learning.}   Explicit preferences focus on the distinction between the preferred response \( y_w \) and each dispreferred response \( y_{lj} \). To quantify this, we compute the loss iteratively for \( y_w \) and all \( k \) dispreferred outputs:
\begin{equation}
\label{DPO_external}
\resizebox{0.5\textwidth}{!}{$
\begin{aligned}
    L_{\text{E}} = &- \sum_{j=1}^{k} \mathbb{E}_{(x, y_w, y_{lj}) \sim D} \left[ \log \sigma \left( \gamma \log \frac{\pi_\theta(y_w|x)}{\pi_{\text{sft}}(y_w|x)} \right. \right. \\
    &\quad \left. \left. - \gamma \log \frac{\pi_\theta(y_{lj}|x)}{\pi_{\text{sft}}(y_{lj}|x)} \right) \right].
\end{aligned}
$}
\end{equation}
Here, \( L_{\text{E}} \) encourages the model to assign higher probabilities to the preferred response \( y_w \) compared to each dispreferred response \( y_{lj} \). The term \( \gamma \) acts as a temperature parameter, controlling the strength of the preference signal. By iterating over all the dispreferred responses, the model learns to explicitly prioritize \( y_w \) over suboptimal alternatives.

\noindent\textbf{Implicit Preference Learning.} Implicit preferences focus on the relative quality among dispreferred responses (e.g., high-ranked vs. low-ranked). To capture these nuanced differences, we compute the loss for  all dispreferred pairs \( (y_{lj}, y_{lm}) \):
\begin{equation}
\label{DPO_internal}
\resizebox{0.5\textwidth}{!}{$
\begin{aligned}
    L_{\text{I}} = &- \sum_{j=1}^{k} \sum_{m=j+1}^{k} \mathbb{E}_{(x, y_{lj}, y_{lm}) \sim D} \left[ \log \sigma \left( \gamma \log \frac{\pi_\theta(y_{lj}|x)}{\pi_{\text{sft}}(y_{lj}|x)} \right. \right. \\
    &\quad \left. \left. - \gamma \log \frac{\pi_\theta(y_{lm}|x)}{\pi_{\text{sft}}(y_{lm}|x)} \right) \right].
\end{aligned}
$}
\end{equation}
Here, \( L_{\text{I}} \) encourages the model to distinguish between dispreferred responses based on their relative quality. By comparing all possible pairs \( (y_{lj}, y_{lm}) \), the model learns to implicitly rank dispreferred responses, ensuring that higher-quality but still suboptimal responses are prioritized over lower-quality ones.
The total loss is defined as the sum of the explicit and implicit losses:
\begin{equation}
L_{\text{HSCR}} = L_{\text{E}} + L_{\text{I}}.
\end{equation}
This loss function enables the model to simultaneously capture both explicit and implicit preferences. By integrating these two levels of preference learning, our HSCR effectively captures both high-level distinctions and fine-grained nuances. 

%This multi-level optimization not only enhances modality alignment but also improves the model's trustworthiness.
\begin{table*}[ht!]
\centering
\renewcommand{\arraystretch}{0.8}
\resizebox{\textwidth}{!}{
\begin{tabular}{llcc|cc|cc}  
\toprule
\textbf{Method} & & \multicolumn{2}{c|}{\cellcolor[HTML]{E8F5E9}\textbf{RAD-VQA}} & \multicolumn{2}{c|}{\cellcolor[HTML]{FCE4EC}\textbf{SLAKE}} & \multicolumn{2}{c}{\cellcolor[HTML]{FFF9E3}\textbf{PathVQA}} \\
 & & \cellcolor[HTML]{E8F5E9}\textbf{Open} & \cellcolor[HTML]{E8F5E9}\textbf{Closed} & \cellcolor[HTML]{FCE4EC}\textbf{Open} & \cellcolor[HTML]{FCE4EC}\textbf{Closed} & \cellcolor[HTML]{FFF9E3}\textbf{Open} & \cellcolor[HTML]{FFF9E3}\textbf{Closed} \\
\midrule
\rowcolor[HTML]{F3F4F6}
\multicolumn{8}{l}{\textit{Representative \& SoTA methods reported in the literature (Non-VLMs Based Methods)}} \\
\midrule
VL Encoder--Decoder~\citep{bazi2023vision} & & \cellcolor[HTML]{E8F5E9}- & \cellcolor[HTML]{E8F5E9}82.47 & \cellcolor[HTML]{FCE4EC}- & \cellcolor[HTML]{FCE4EC}- & \cellcolor[HTML]{FFF9E3}- & \cellcolor[HTML]{FFF9E3}85.61 \\
Q2ATransformer~\citep{liu2023q2atransformer} & & \cellcolor[HTML]{E8F5E9}- & \cellcolor[HTML]{E8F5E9}81.20 & \cellcolor[HTML]{FCE4EC}- & \cellcolor[HTML]{FCE4EC}- & \cellcolor[HTML]{FFF9E3}54.85 & \cellcolor[HTML]{FFF9E3}88.85 \\
Prefix T. Medical LM~\citep{van2023open} & & \cellcolor[HTML]{E8F5E9}- & \cellcolor[HTML]{E8F5E9}- & \cellcolor[HTML]{FCE4EC}- & \cellcolor[HTML]{FCE4EC}82.01 & \cellcolor[HTML]{FFF9E3}- & \cellcolor[HTML]{FFF9E3}87.00 \\
PubMedCLIP~\citep{eslami2023pubmedclip} & & \cellcolor[HTML]{E8F5E9}- & \cellcolor[HTML]{E8F5E9}80.00 & \cellcolor[HTML]{FCE4EC}- & \cellcolor[HTML]{FCE4EC}82.50 & \cellcolor[HTML]{FFF9E3}- & \cellcolor[HTML]{FFF9E3}- \\
BiomedCLIP~\citep{zhang2023large} & & \cellcolor[HTML]{E8F5E9}- & \cellcolor[HTML]{E8F5E9}79.80 & \cellcolor[HTML]{FCE4EC}- & \cellcolor[HTML]{FCE4EC}89.70 & \cellcolor[HTML]{FFF9E3}- & \cellcolor[HTML]{FFF9E3}- \\
M2I2~\citep{li2022self} & & \cellcolor[HTML]{E8F5E9}- & \cellcolor[HTML]{E8F5E9}83.50 & \cellcolor[HTML]{FCE4EC}- & \cellcolor[HTML]{FCE4EC}91.10 & \cellcolor[HTML]{FFF9E3}- & \cellcolor[HTML]{FFF9E3}88.00 \\
BiomedGPT-S~\citep{zhang2023biomedgpt} & & \cellcolor[HTML]{E8F5E9}13.40 & \cellcolor[HTML]{E8F5E9}57.80  & \cellcolor[HTML]{FCE4EC}66.50 & \cellcolor[HTML]{FCE4EC}73.30 & \cellcolor[HTML]{FFF9E3}10.70 & \cellcolor[HTML]{FFF9E3}84.20 \\
BiomedGPT-M~\citep{zhang2023biomedgpt} & & \cellcolor[HTML]{E8F5E9}53.60 & \cellcolor[HTML]{E8F5E9}65.07 & \cellcolor[HTML]{FCE4EC}78.30 & \cellcolor[HTML]{FCE4EC}86.80 & \cellcolor[HTML]{FFF9E3}12.5 & \cellcolor[HTML]{FFF9E3}85.70 \\
CLIP-ViT w/ GPT2-XL~\citep{radford2021learning} & & \cellcolor[HTML]{E8F5E9}- & \cellcolor[HTML]{E8F5E9}- & \cellcolor[HTML]{FCE4EC}84.30 & \cellcolor[HTML]{FCE4EC}82.10 & \cellcolor[HTML]{FFF9E3}40.0 & \cellcolor[HTML]{FFF9E3}87.00 \\
\midrule
\rowcolor[HTML]{F3F4F6}
\multicolumn{8}{l}{\textit{Zero-shot results}} \\
\midrule
GPT-4o~\citep{hurst2024gpt} & & \cellcolor[HTML]{E8F5E9}{51.6} & \cellcolor[HTML]{E8F5E9}{63.97} & \cellcolor[HTML]{FCE4EC}59.06 & \cellcolor[HTML]{FCE4EC}71.63 & \cellcolor[HTML]{FFF9E3}24.14 & \cellcolor[HTML]{FFF9E3}75.97 \\
LLaVA1.5~\citep{Liu_2024_CVPR} & & \cellcolor[HTML]{E8F5E9}{23.63} & \cellcolor[HTML]{E8F5E9}{50.74} & \cellcolor[HTML]{FCE4EC}35.23 & \cellcolor[HTML]{FCE4EC}52.16 & \cellcolor[HTML]{FFF9E3}11.85 & \cellcolor[HTML]{FFF9E3}52.76 \\
Med-Flamingo~\citep{moor2023med} & & \cellcolor[HTML]{E8F5E9}10.32 & \cellcolor[HTML]{E8F5E9}52.21 & \cellcolor[HTML]{FCE4EC}{8.46} & \cellcolor[HTML]{FCE4EC}37.02 & \cellcolor[HTML]{FFF9E3}1.23 & \cellcolor[HTML]{FFF9E3}45.59 \\
PMC-VQA~\citep{zhang2023pmc} & & \cellcolor[HTML]{E8F5E9}6.26 & \cellcolor[HTML]{E8F5E9}41.54 & \cellcolor[HTML]{FCE4EC}7.29 & \cellcolor[HTML]{FCE4EC}33.89 & \cellcolor[HTML]{FFF9E3}{1.02} & \cellcolor[HTML]{FFF9E3}40.10 \\
SQ-LLaVA~\citep{sun2025sq} & & \cellcolor[HTML]{E8F5E9}23.91 & \cellcolor[HTML]{E8F5E9}52.57 & \cellcolor[HTML]{FCE4EC}{40.04} & \cellcolor[HTML]{FCE4EC}{57.45} & \cellcolor[HTML]{FFF9E3}\underline{12.24} & \cellcolor[HTML]{FFF9E3}53.79 \\
ST-LLaVA~\citep{sun2024stllava} & & \cellcolor[HTML]{E8F5E9}\underline{33.81} & \cellcolor[HTML]{E8F5E9}\underline{59.16} & \cellcolor[HTML]{FCE4EC}{40.13} & \cellcolor[HTML]{FCE4EC}{55.53} & \cellcolor[HTML]{FFF9E3}10.38 & \cellcolor[HTML]{FFF9E3}52.05 \\
VCD~\citep{leng2024mitigating} & & \cellcolor[HTML]{E8F5E9}30.54 & \cellcolor[HTML]{E8F5E9}55.88 & \cellcolor[HTML]{FCE4EC}{42.92} & \cellcolor[HTML]{FCE4EC}56.93 & \cellcolor[HTML]{FFF9E3}{9.13} & \cellcolor[HTML]{FFF9E3}{58.16} \\
LiPO~\citep{liu2024lipo}& & \cellcolor[HTML]{E8F5E9}31.85 & \cellcolor[HTML]{E8F5E9}57.37& \cellcolor[HTML]{FCE4EC}\underline{43.18} & \cellcolor[HTML]{FCE4EC}\underline{58.13} & \cellcolor[HTML]{FFF9E3}{9.37} & \cellcolor[HTML]{FFF9E3}\underline{60.17} \\
LLaVA-Med1.5~\citep{li2024llava}& & \cellcolor[HTML]{E8F5E9}32.31 & \cellcolor[HTML]{E8F5E9}56.62& \cellcolor[HTML]{FCE4EC}{42.45} & \cellcolor[HTML]{FCE4EC}56.49 & \cellcolor[HTML]{FFF9E3}{10.01} & \cellcolor[HTML]{FFF9E3}{59.75} \\
HSCR (Ours) & & \cellcolor[HTML]{E8F5E9}\textbf{35.92(+3.61)} & \cellcolor[HTML]{E8F5E9}\textbf{60.13(+3.51)} & \cellcolor[HTML]{FCE4EC}\textbf{45.32(+2.87)} & \cellcolor[HTML]{FCE4EC}\textbf{63.46(+6.97)} & \cellcolor[HTML]{FFF9E3}\textbf{12.36(+2.35)} & \cellcolor[HTML]{FFF9E3}\textbf{64.17(+4.42)} \\
\bottomrule
\end{tabular}
}
\caption{Performance on Med-VQA tasks. \textbf{Bold} denotes the best performance,\underline{underlined} denotes the second-best.}
\label{tab:main_table}
\end{table*}

\begin{table*}[ht]
    \centering
    \renewcommand{\arraystretch}{0.8} % 调整行高
    \resizebox{\textwidth}{!}{%
    \begin{tabular}{@{}l|cc|ccccc|c@{}}
        \toprule
        \multirow{2}{*} & \multicolumn{2}{c|}{\cellcolor[HTML]{E8F5E9}Question Types} & \multicolumn{5}{c|}{\cellcolor[HTML]{FCE4EC}Domains} & \cellcolor[HTML]{FFF9E3}Overall \\
        & \cellcolor[HTML]{E8F5E9}Conversation & \cellcolor[HTML]{E8F5E9}Description & \cellcolor[HTML]{FCE4EC}CXR & \cellcolor[HTML]{FCE4EC}MRI & \cellcolor[HTML]{FCE4EC}Histology & \cellcolor[HTML]{FCE4EC}Gross & \cellcolor[HTML]{FCE4EC}CT & \cellcolor[HTML]{FFF9E3} \\
        \midrule
        LLaVA & \cellcolor[HTML]{E8F5E9}39.4 & \cellcolor[HTML]{E8F5E9}26.2 & \cellcolor[HTML]{FCE4EC}41.6 & \cellcolor[HTML]{FCE4EC}33.4 & \cellcolor[HTML]{FCE4EC}38.4 & \cellcolor[HTML]{FCE4EC}32.9 & \cellcolor[HTML]{FCE4EC}33.4 & \cellcolor[HTML]{FFF9E3}36.1 \\
        \midrule
        LLaVA-Med & \cellcolor[HTML]{E8F5E9} & \cellcolor[HTML]{E8F5E9} & \cellcolor[HTML]{FCE4EC} & \cellcolor[HTML]{FCE4EC} & \cellcolor[HTML]{FCE4EC} & \cellcolor[HTML]{FCE4EC} & \cellcolor[HTML]{FCE4EC} & \cellcolor[HTML]{FFF9E3} \\
        Pretrain & \cellcolor[HTML]{E8F5E9}22.6 & \cellcolor[HTML]{E8F5E9}25.2 & \cellcolor[HTML]{FCE4EC}25.8 & \cellcolor[HTML]{FCE4EC}19.0 & \cellcolor[HTML]{FCE4EC}24.8 & \cellcolor[HTML]{FCE4EC}24.7 & \cellcolor[HTML]{FCE4EC}22.2 & \cellcolor[HTML]{FFF9E3}23.3 \\
        SFT (10K) & \cellcolor[HTML]{E8F5E9}42.4 & \cellcolor[HTML]{E8F5E9}32.5 & \cellcolor[HTML]{FCE4EC}46.1 & \cellcolor[HTML]{FCE4EC}36.7 & \cellcolor[HTML]{FCE4EC}43.5 & \cellcolor[HTML]{FCE4EC}34.7 & \cellcolor[HTML]{FCE4EC}37.5 & \cellcolor[HTML]{FFF9E3}39.9 \\
        SFT (60K) & \cellcolor[HTML]{E8F5E9}53.7 & \cellcolor[HTML]{E8F5E9}36.9 & \cellcolor[HTML]{FCE4EC}57.3 & \cellcolor[HTML]{FCE4EC}39.8 & \cellcolor[HTML]{FCE4EC}49.8 & \cellcolor[HTML]{FCE4EC}47.4 & \cellcolor[HTML]{FCE4EC}52.4 & \cellcolor[HTML]{FFF9E3}49.4 \\
        SFT (60K-IM) & \cellcolor[HTML]{E8F5E9}55.1 & \cellcolor[HTML]{E8F5E9}36.4 & \cellcolor[HTML]{FCE4EC}56.2 & \cellcolor[HTML]{FCE4EC}40.4 & \cellcolor[HTML]{FCE4EC}52.7 & \cellcolor[HTML]{FCE4EC}51.8 & \cellcolor[HTML]{FCE4EC}50.1 & \cellcolor[HTML]{FFF9E3}50.2 \\
        \midrule
        LLaVA-Med1.5 & \cellcolor[HTML]{E8F5E9} & \cellcolor[HTML]{E8F5E9} & \cellcolor[HTML]{FCE4EC} & \cellcolor[HTML]{FCE4EC} & \cellcolor[HTML]{FCE4EC} & \cellcolor[HTML]{FCE4EC} & \cellcolor[HTML]{FCE4EC} & \cellcolor[HTML]{FFF9E3} \\
        SFT (60K-IM) & \cellcolor[HTML]{E8F5E9}58.6 & \cellcolor[HTML]{E8F5E9}42.5 & \cellcolor[HTML]{FCE4EC}59.6 & \cellcolor[HTML]{FCE4EC}46.5 & \cellcolor[HTML]{FCE4EC}58.8 & \cellcolor[HTML]{FCE4EC}52.8 & \cellcolor[HTML]{FCE4EC}53.8 & \cellcolor[HTML]{FFF9E3}54.4 \\
        \textbf{HSCR (2K)} & \cellcolor[HTML]{E8F5E9}\textbf{59.4 (+0.8)} & \cellcolor[HTML]{E8F5E9}\textbf{52.9 (+10.4)} & \cellcolor[HTML]{FCE4EC}\textbf{62.0 (+2.4)} & \cellcolor[HTML]{FCE4EC}\textbf{47.9 (+1.4)} & \cellcolor[HTML]{FCE4EC}\textbf{65.1 (+6.3)} & \cellcolor[HTML]{FCE4EC}\textbf{53.5 (+0.7)} & \cellcolor[HTML]{FCE4EC}\textbf{59.5 (+5.7)} & \cellcolor[HTML]{FFF9E3}\textbf{57.7 (+3.3)} \\
        \bottomrule
    \end{tabular}
    }
\caption{Performance on captioning and instruction-following tasks. IM denotes visual instruction-following data enhanced with figure references from PubMed Central articles~\citep{pringle2006appropriate}.}

    \label{tab:tab2}
\end{table*}

\section{Experiments}

\subsection{Experiements Setup}  
\noindent\textbf{Training Settings.} Following previous research~\citep{li2024llava}, we adopt CLIP-ViT-L/14@336px~\citep{radford2021learning} as the visual encoder to extract visual features from medical images, Mistral-7B~\citep{jiang2023mistral} as the text encoder, and a two-layer MLP with a GeLU~\citep{hendrycks2016gaussian} activation function as the projector to align the text and visual encoders.

\noindent\textbf{Details of Hyperparameters}
\label{ref:expe_detail}
We sample 2,000 entries from the dataset used in instruction-tuning stage to construct preference datasets as described in Methods~\ref{methods1}. For this process, we set the parameters as follows: \( j = 3 \), \( \beta = 0.9 \), \( n = 10 \), and \( \gamma = 0.1 \). Multi-level preference optimization is performed using the constructed implicit and explicit preference datasets, with LoRA~\citep{hu2021lora} applied at a rank of 16. Training is conducted for 2 epochs with a learning rate of 5e-7 and without weight decay. The training hyperparameters are in Table~\ref{tab:setting}.

\noindent\textbf{Evaluation Datasets.}
We evaluate HSCR on all benchmarks from LLaVA-Med~\citep{li2024llava}, including Rad-VQA~\citep{lau2018dataset}, SLAKE~\citep{liu2021slake}, and PathVQA~\citep{he2020pathvqa}, covering both open-ended and closed-ended settings (details in Appendix~\ref{ref:evaldata}). For captioning and instruction-following tasks, we use LLaVA-Med's evaluation datasets, which include unseen image-caption pairs from PMC-15M~\citep{zhang2023large}, generating conversation and detailed description questions across five medical modalities and multi-turn dialogues. This ensures a thorough evaluation of medical VLMs across diverse domains and tasks.

\noindent\textbf{Evaluation Metrics.}
For Med-VQA tasks, we follow prior work~\citep{sun2024stllava,zhang2023pmc}, using accuracy for closed-set (yes/no) questions and recall for open-set (free-form) questions. For captioning and instruction-following tasks, we employ GPT-4~\citep{gpt4} as an automated evaluator. GPT-4 generates a reference answer based on the image and caption, against which the candidate model’s response is scored. Each response receives an overall score on a 10-point scale, normalized for comparability, along with a detailed explanation justifying the rating.

\noindent \textbf{Baselines.}  
We select a variety of strong baselines. For general-purpose VLMs, we include models such as LLaVA1.5~\citep{Liu_2024_CVPR} and SQ-LLaVA~\citep{sun2025sq}, as they represent top-performing VLMs trained on general-domain instruction-following datasets without incorporating medical data.  
For medical-specific VLMs, we select leading models, including LLaVA-Med1.5~\citep{li2024llava}, Med-Flamingo~\citep{moor2023med}, and PMC-VQA~\citep{zhang2023pmc}, which are specifically trained on medical instruction-following datasets and demonstrate superior performance on medical tasks.  
Additionally, for methods aimed at enhancing the alignment of medical VLMs, we include the latest approaches such as ST-LLaVA~\citep{sun2024stllava}, VCD~\citep{leng2024mitigating} and LiPO~\citep{liu2024lipo}.

\begin{table*}[!t]
\centering
\renewcommand{\arraystretch}{1}
\resizebox{\textwidth}{!}{
\begin{tabular}{llcc|cc|cc}
\toprule
\textbf{Method} & & \multicolumn{2}{c|}{\cellcolor[HTML]{E8F5E9}\textbf{RAD-VQA}} & \multicolumn{2}{c|}{\cellcolor[HTML]{FCE4EC}\textbf{SLAKE}} & \multicolumn{2}{c}{\cellcolor[HTML]{FFF9E3}\textbf{PathVQA}} \\
 & & \cellcolor[HTML]{E8F5E9}\textbf{Open} & \cellcolor[HTML]{E8F5E9}\textbf{Closed} & \cellcolor[HTML]{FCE4EC}\textbf{Open} & \cellcolor[HTML]{FCE4EC}\textbf{Closed} & \cellcolor[HTML]{FFF9E3}\textbf{Open} & \cellcolor[HTML]{FFF9E3}\textbf{Closed} \\
\midrule
LLaVA-Med1.5~\citep{li2024llava} & & \cellcolor[HTML]{E8F5E9}32.31 & \cellcolor[HTML]{E8F5E9}56.62 & \cellcolor[HTML]{FCE4EC}\underline{42.45} & \cellcolor[HTML]{FCE4EC}56.49 & \cellcolor[HTML]{FFF9E3}\underline{10.01} & \cellcolor[HTML]{FFF9E3}{59.75} \\
GPT-4o Generated~\citep{hurst2024gpt} & & \cellcolor[HTML]{E8F5E9}\underline{33.14 (+0.83)} & \cellcolor[HTML]{E8F5E9}\underline{57.20 (+0.58)} & \cellcolor[HTML]{FCE4EC}41.83 (-0.62) & \cellcolor[HTML]{FCE4EC}\underline{57.96 (+1.47)} & \cellcolor[HTML]{FFF9E3}9.55 (-0.46) & \cellcolor[HTML]{FFF9E3}\underline{60.35 (+0.60)} \\
HSCR (Ours) & & \cellcolor[HTML]{E8F5E9}\textbf{35.92 (+3.61)} & \cellcolor[HTML]{E8F5E9}\textbf{60.13 (+3.51)} & \cellcolor[HTML]{FCE4EC}\textbf{45.32 (+2.87)} & \cellcolor[HTML]{FCE4EC}\textbf{63.46 (+6.97)} & \cellcolor[HTML]{FFF9E3}\textbf{12.36 (+2.35)} & \cellcolor[HTML]{FFF9E3}\textbf{64.17 (+4.42)} \\
\bottomrule
\end{tabular}
}
\caption{Ablation of preference data construction.}
\label{tab:data_ab}
\end{table*}
\begin{table*}[!ht]
\centering
\resizebox{1\textwidth}{!}{%
\begin{tabular}{cc|cc|cc|cc}
\toprule
\textbf{Explicit Pref.} & \textbf{Implicit Pref.} & \multicolumn{2}{c|}{\cellcolor[HTML]{E8F5E9}\textbf{RAD-VQA}} & \multicolumn{2}{c|}{\cellcolor[HTML]{FCE4EC}\textbf{SLAKE}} & \multicolumn{2}{c}{\cellcolor[HTML]{FFF9E3}\textbf{PathVQA}} \\
 & & \cellcolor[HTML]{E8F5E9}\textbf{Open} & \cellcolor[HTML]{E8F5E9}\textbf{Closed} & \cellcolor[HTML]{FCE4EC}\textbf{Open} & \cellcolor[HTML]{FCE4EC}\textbf{Closed} & \cellcolor[HTML]{FFF9E3}\textbf{Open} & \cellcolor[HTML]{FFF9E3}\textbf{Closed} \\
\midrule
$\times$ & $\times$ & \cellcolor[HTML]{E8F5E9}32.31 & \cellcolor[HTML]{E8F5E9}56.62 & \cellcolor[HTML]{FCE4EC}{42.45} & \cellcolor[HTML]{FCE4EC}56.49 & \cellcolor[HTML]{FFF9E3}{10.01} & \cellcolor[HTML]{FFF9E3}59.75 \\
$\checkmark$ & $\times$ & \cellcolor[HTML]{E8F5E9}33.69 (+1.38) & \cellcolor[HTML]{E8F5E9}\underline{58.31 (+1.69)} & \cellcolor[HTML]{FCE4EC}43.14 (+0.69) & \cellcolor[HTML]{FCE4EC}57.78 (+1.29) & \cellcolor[HTML]{FFF9E3}10.46 (+0.45) & \cellcolor[HTML]{FFF9E3}60.05 (+0.30) \\ 
$\times$ & $\checkmark$ & \cellcolor[HTML]{E8F5E9}\underline{34.13 (+1.82)}&57.65 (+1.03) \cellcolor[HTML]{E8F5E9} & \cellcolor[HTML]{FCE4EC}\underline{43.95 (+1.50)} & \cellcolor[HTML]{FCE4EC}\underline{60.32 (+3.83)} & \cellcolor[HTML]{FFF9E3}\underline{11.24 (+1.23)} & \cellcolor[HTML]{FFF9E3}\underline{62.12 (+2.37)} \\ 
$\checkmark$ & $\checkmark$ & \cellcolor[HTML]{E8F5E9}\textbf{35.92 (+3.61)} & \cellcolor[HTML]{E8F5E9}\textbf{60.13 (+3.51)} & \cellcolor[HTML]{FCE4EC}\textbf{45.32 (+2.87)} & \cellcolor[HTML]{FCE4EC}\textbf{63.46 (+6.97)} & \cellcolor[HTML]{FFF9E3}\textbf{12.36 (+2.35)} & \cellcolor[HTML]{FFF9E3}\textbf{64.17 (+4.42)} \\
\bottomrule
\end{tabular}%
}
\caption{Ablation study on the impact of explicit and implicit preferences.}
\label{tab:pref_ab}
\end{table*}
\begin{table*}[!ht]
\centering
\resizebox{1\textwidth}{!}{%
\begin{tabular}{cc|cc|cc|cc}
\toprule
\textbf{Mask Strategy} & & \multicolumn{2}{c|}{\cellcolor[HTML]{E8F5E9}\textbf{RAD-VQA}} & \multicolumn{2}{c|}{\cellcolor[HTML]{FCE4EC}\textbf{SLAKE}} & \multicolumn{2}{c}{\cellcolor[HTML]{FFF9E3}\textbf{PathVQA}} \\
 & & \cellcolor[HTML]{E8F5E9}\textbf{Open} & \cellcolor[HTML]{E8F5E9}\textbf{Closed} & \cellcolor[HTML]{FCE4EC}\textbf{Open} & \cellcolor[HTML]{FCE4EC}\textbf{Closed} & \cellcolor[HTML]{FFF9E3}\textbf{Open} & \cellcolor[HTML]{FFF9E3}\textbf{Closed} \\
\midrule
Baseline & & \cellcolor[HTML]{E8F5E9}32.31 & \cellcolor[HTML]{E8F5E9}56.62 & \cellcolor[HTML]{FCE4EC}42.45 & \cellcolor[HTML]{FCE4EC}56.49 & \cellcolor[HTML]{FFF9E3}10.01 & \cellcolor[HTML]{FFF9E3}59.75 \\
Pixel-Level Mask & & \cellcolor[HTML]{E8F5E9}33.15 (+0.84) & \cellcolor[HTML]{E8F5E9}57.17 (+0.55) & \cellcolor[HTML]{FCE4EC}42.65 (+0.20) & \cellcolor[HTML]{FCE4EC}57.49 (+1.00) & \cellcolor[HTML]{FFF9E3}10.45 (+0.44) & \cellcolor[HTML]{FFF9E3}60.79 (+1.04) \\
Patch-Level Mask & & \cellcolor[HTML]{E8F5E9}\underline{34.03 (+1.72)} & \cellcolor[HTML]{E8F5E9}57.91 (+1.29) & \cellcolor[HTML]{FCE4EC}43.13 (+0.68) & \cellcolor[HTML]{FCE4EC}58.44 (+1.95) & \cellcolor[HTML]{FFF9E3}11.01 (+1.00) & \cellcolor[HTML]{FFF9E3}61.83 (+2.08) \\
Latent Space Mask & & \cellcolor[HTML]{E8F5E9}33.45 (+1.14) & \cellcolor[HTML]{E8F5E9}\underline{58.79 (+2.17)} & \cellcolor[HTML]{FCE4EC}\underline{43.56 (+1.11)} & \cellcolor[HTML]{FCE4EC}\underline{60.32 (+3.83)} & \cellcolor[HTML]{FFF9E3}\underline{11.23 (+1.22)} & \cellcolor[HTML]{FFF9E3}\underline{62.77 (+3.02)} \\
Visual Token Dropout & & \cellcolor[HTML]{E8F5E9}\textbf{35.92 (+3.61)} & \cellcolor[HTML]{E8F5E9}\textbf{60.13 (+3.51)} & \cellcolor[HTML]{FCE4EC}\textbf{45.32 (+2.87)} & \cellcolor[HTML]{FCE4EC}\textbf{63.46 (+6.97)} & \cellcolor[HTML]{FFF9E3}\textbf{12.36 (+2.35)} & \cellcolor[HTML]{FFF9E3}\textbf{64.17 (+4.42)} \\
\bottomrule
\end{tabular}%
}
\caption{Performance comparison of different masking strategies.}
\label{tab:mask_methods}
\end{table*}

\subsection{Main Results}
\noindent \textbf{Performance on Med-VQA Tasks.} As shown in Table~\ref{tab:main_table}, our method achieves SOTA performance in the zero-shot setting among open-source models. Specifically, in the closed setting, our method performs nearly on par with GPT-4 on RAD-VQA and achieves a notable 7\% performance improvement over LLaVA-Med1.5 on the SLAKE dataset. More importantly, for challenging open-set questions, our approach consistently demonstrates performance gains with only 2,000 training entries (See Appendix~\ref{ref:lipo} for more discussion). 

\noindent\textbf{Performance on Captioning and Instruction-Following Tasks.} As demonstrated in Table~\ref{tab:tab2}, our method achieves promising improvements in both captioning and instruction-following capabilities for medical VLMs. It enhances conversational and descriptive performance across diverse medical modalities, consistently surpassing LLaVA-Med1.5. Notably, with only 2,000 training samples, our approach delivers more substantial performance gains compared to scaling supervised fine-tuning (SFT) data from 10k to 50k entries, achieving a 10.4\% improvement in captioning accuracy versus a 4.4\% gain from data expansion.  Furthermore, our method demonstrates consistent performance improvements across all five medical modalities, highlighting its robustness and adaptability. This not only elevates overall performance but also enhances the trustworthiness and reliability of the model in real-world medical applications.
\section{Ablation and Analysis}

\begin{figure*}[ht!]
    \centering
    \includegraphics[width=1\linewidth]{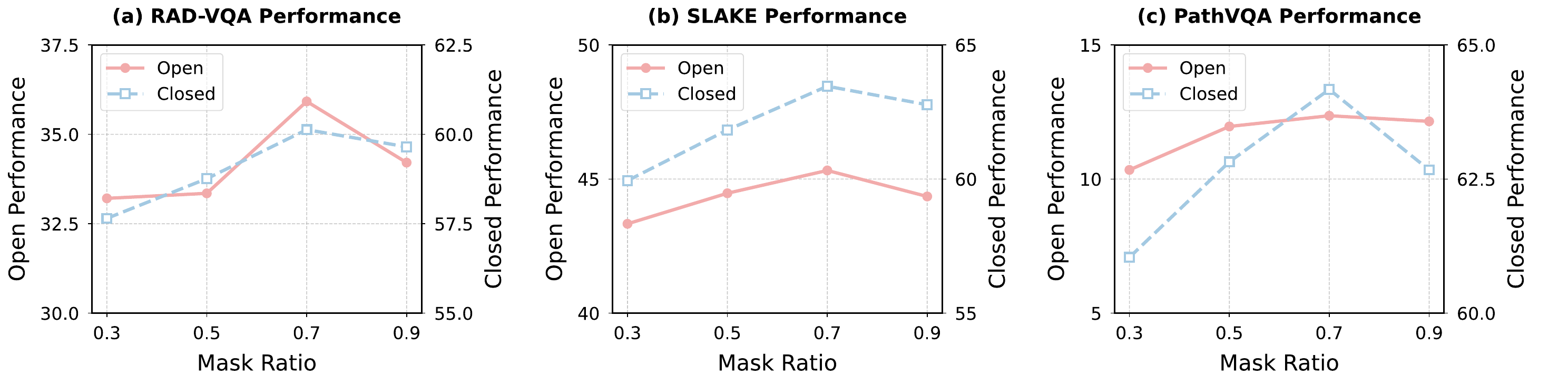}
\caption{Performance comparison with different mask ratios. Details in Appendix Table~\ref{tab:mask_ratio}. }
    \label{fig:mask_ratio}
\end{figure*}

\begin{figure*}[ht!]
    \centering
    \includegraphics[width=1\linewidth]{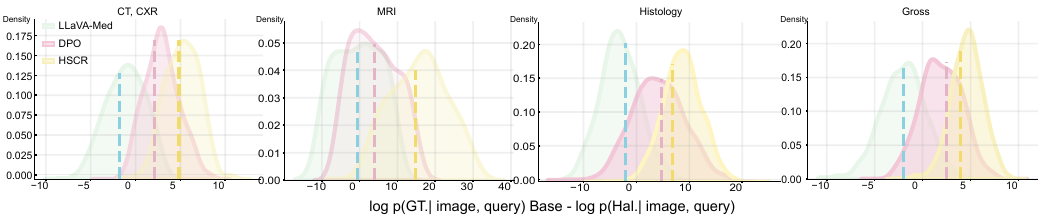}
\caption{The x-axis represents $\log p(\mathrm{Ground\;Truth\;Response}) - \log p(\mathrm{Hallucinatory})$ across five different medical modalities. The dashed line in the figure indicates the median value. A larger value signifies a stronger ability of the VLMs to distinguish hallucinatory responses.}
% The colors correspond to three different methods: \textcolor[HTML]{e8f5e9}{\rule{1em}{1.5ex}} for the LLaVA-Med model, \textcolor[HTML]{fce4ec}{\rule{1em}{1.5ex}} for DPO, and \textcolor[HTML]{fff9e3}{\rule{1em}{1.5ex}} for our HSCR.
    \label{fig:log}
\end{figure*}

\noindent \textbf{Ablation on Preference Dataset Construction.}
We evaluate our token-level preference datasets against GPT-4o-generated preferences~\citep{hurst2024gpt} in a binary setting (Table~\ref{tab:data_ab}). While GPT-4o preferences yield modest gains in closed-ended tasks (+1.47\% on SLAKE, +0.58\% on RAD-VQA), they degrade open-ended performance (-0.62\% on SLAKE, -0.46\% on PathVQA). This highlights the limitations of external preference reward signals from stronger models in guiding medical VLMs, particularly for complex open-ended queries. The key issue stems from the mismatch between external preferences and inherent VLM misalignment biases, leading to suboptimal optimization.

In contrast, HSCR constructs preference datasets by exposing and leveraging misalignment responses inherent to medical VLMs. By precisely correcting these intrinsic misalignment behaviors, HSCR significantly enhances both the performance and trustworthiness of medical VLMs.
%This approach delivers substantial improvements: +3.61\% (open) and +3.51\% (closed) on RAD-VQA; +2.87\% and +6.97\% on SLAKE; +2.35\% and +4.42\% on PathVQA.

\noindent \textbf{Ablation on Implicit and Explicit Preferences.}
As shown in Table~\ref{tab:pref_ab}, we analyze the effects of explicit and implicit preferences on medical VLMs. Implicit preferences outperform explicit ones by leveraging fine-grained reward gradients to provide nuanced, context-aware guidance, achieving significant gains: 1.82\% on RAD-VQA, 3.83\% on SLAKE, and 2.37\% on PathVQA. In contrast, explicit preferences, while beneficial, are more effective for high-level, coarse-grained optimization, resulting in comparatively modest improvements. The combination of both approaches yields optimal results, with improvements of up to 3.61\% on RAD-VQA, 6.97\% on SLAKE, and 4.42\% on PathVQA. This demonstrates their complementary roles: \textit{explicit preferences offer broad, high-level direction, while implicit preferences capture subtle, task-specific signals}, enhancing the robustness and trustworthiness of medical VLMs.

\begin{figure*}[!ht]
    \centering
    \includegraphics[width=1\linewidth]{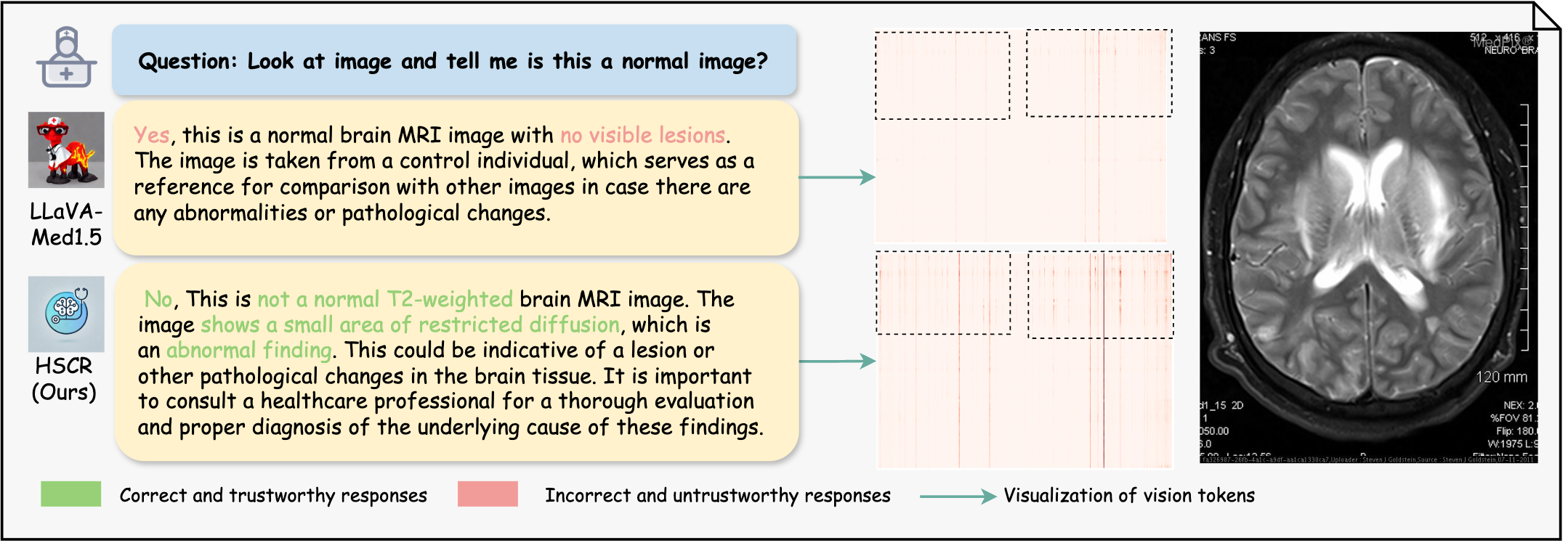}
\caption{Zero-shot comparison of LLaVA-Med1.5~\citep{li2024llava} and HSCR. HSCR generates more accurate, detailed, and image-relevant responses, with attention maps (highlighted in black) showing stronger focus on the image modality, achieving superior alignment and trustworthiness. Additional cases are in Appendix~\ref{fig:more_case}.}
   \label{fig:case}
\end{figure*}

\noindent \textbf{Ablation on Mask Ratio.} We conducted an ablation study to investigate the impact of different mask ratios on model performance in Figure~\ref{fig:mask_ratio}. Our findings indicate that the proposed method exhibits robustness to variations in the mask ratio, consistently outperforming the baseline across all tested ratios. Notably, the performance gains are more substantial for mask ratios exceeding 0.5. This phenomenon can be attributed to the necessity of a sufficiently high mask ratio to effectively disrupt visual information, thereby eliciting meaningful misalignment responses. Based on our empirical results, we adopted a mask ratio of 0.7 for our experiments, as it consistently yields optimal performance across a wide range of scenarios.
\begin{figure*}[ht!]
    \centering
    \includegraphics[width=1\linewidth]{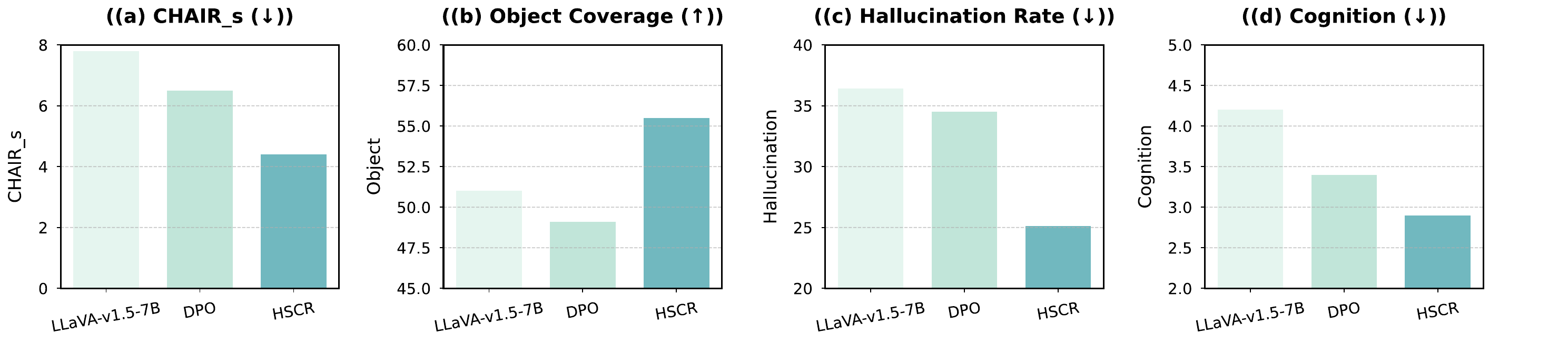}
\caption{Performance on general multimodal tasks. Details in Appendix Table~\ref{tab:app_general}. }
    \label{fig:general}
\end{figure*}
\begin{figure}[!ht]
    \centering
    \includegraphics[width=1\linewidth]{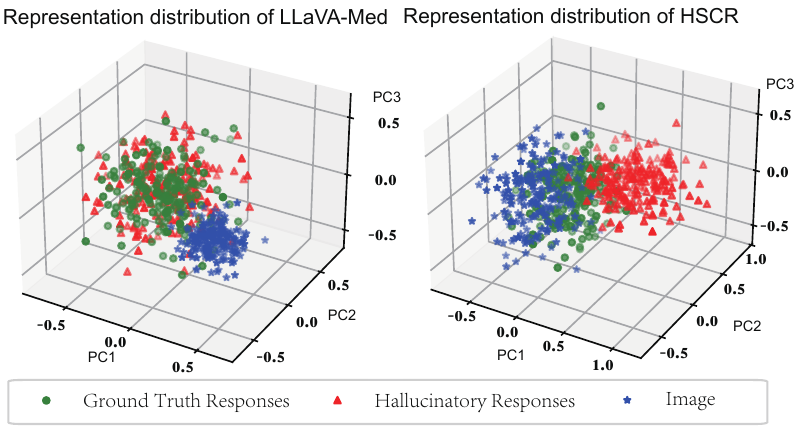}
\caption{Representation distribution comparison.}
   \label{fig:3d}
\end{figure}

\noindent \textbf{Effect of Mask Strategies.} In this ablation study, we fixed the mask ratio at the optimal value of 70\% and systematically evaluated various levels of mask application (see Appendix~\ref{ref:mask} for detailed implementation). As shown in Table~\ref{tab:mask_methods}, While all four disruption methods demonstrate performance improvements over the baseline, visual token dropout emerges as the most effective strategy. We observe a positive correlation between the proximity of the disruption operation to the LLM backbone's input and the overall model performance. This trend can be explained by the fact that visual token dropout directly eliminates portions of visual information before it reaches the LLM, thereby more effectively triggering the inherent misalignment mechanisms of the underlying language model.

\noindent \textbf{How does HSCR improve modality alignment?}
As illustrated in Figure~\ref{fig:case}, HSCR significantly enhances modality alignment by generating more accurate and detailed responses that are strongly relevant to the image content. In the example from RAD-VQA, HSCR correctly identifies the brain MRI image as normal, providing a precise and trustworthy explanation, while LLaVA-Med1.5 incorrectly identifies an abnormality. The attention map~\citep{vaswani2017attention} of vision tokens, highlighted by the black box, demonstrates that HSCR places greater focus on the image modality, ensuring that the generated responses are well-grounded in visual evidence. This improved attention to image details allows HSCR to achieve superior modality alignment, resulting in more reliable and high-quality answers. By effectively bridging the gap between visual and textual modalities, HSCR not only enhances the accuracy but also boosts the overall trustworthiness of VLMs in medical applications.

\noindent \textbf{HSCR Enhances Representation Learning in Medical VLMs.} Figure~\ref{fig:3d} illustrates the comparative analysis of sample embeddings in Rad-VQA, showcasing correct responses, hallucinated responses, and corresponding image embeddings. The visualization reveals two critical insights: First, the baseline LLaVA-Med model demonstrates significant limitations in distinguishing between correct and hallucinated responses, accompanied by a pronounced modality gap between text and image representations in the latent space. This is evidenced by the substantial distance between embeddings of different modalities. Second, after incorporating HSCR, we observe a notable improvement in the alignment between correct response embeddings and image embeddings, with these representations converging more closely in the latent space. This enhanced clustering indicates improved modality alignment and demonstrates the model's increased capability to discriminate between factual and hallucinated responses. These findings collectively suggest that HSCR effectively augments the representation learning capacity of medical VLMs, leading to strengthened cross-modal alignment and enhanced model trustworthiness.

\noindent \textbf{Generalizability of HSCR to General Multimodal Tasks}
To assess the generalizability of our approach beyond medical domains, we integrated HSCR with the general-purpose VLM LLaVA-v1.5~\citep{liu2024improved} and conducted preference optimization. The model was subsequently evaluated on the widely-used AMBER benchmark~\citep{wang2023llm} for general multimodal tasks (implementation details provided in Appendix~\ref{ref:fig7}). As illustrated in Figure~\ref{fig:general}, HSCR demonstrates significant effectiveness in enhancing model trustworthiness and modality alignment. Notably, our method outperforms the current mainstream approach DPO, indicating its superior capability in mitigating hallucination issues. These results substantiate that HSCR is not only effective for medical VLMs but also demonstrates robust performance in general-domain tasks.

\section{Related Work}
\noindent\textbf{Preference Optimization in VLMs.}
Vision Language Models (VLMs) have achieved significant success in a wide range of tasks~\citep{jiang2024joint,jiang2024moe,liu2024medcot,liu2023deep,liu2025kpl,li2024llava-next}.
Preference optimization has emerged as a critical technique for enhancing the trustworthiness of VLMs. Existing approaches can be broadly categorized into two paradigms: manually constructed preference pairs and automatically generated ones using MLLMs. Methods like RLHF-V~\citep{yu2024rlhf,sun2023aligning} utilize human feedback to refine hallucinated captions, transforming rejected responses into preferred ones. Meanwhile, POVID~\citep{zhou2024aligning,pi2024strengthening} introduces diffusion noise to images, enabling MLLMs to autonomously generate hallucinated content as rejected responses, eliminating the need for human intervention. AMP~\citep{zhang2024automated} enhances stability by producing multiple candidate responses, and RLAIF-V~\citep{yu2024rlaif} improves DPO by aggregating high-quality responses from multiple MLLMs. However, these methods are resource-intensive and have not been validated in resource-constrained clinical settings.

In the medical field, recent concurrent works have also begun exploring preference optimization for alignment. For instance, ST-LLaVA~\citep{sun2024stllava} employs a self-training~\citep{rosenberg2005semi} paradigm to generate preference data and uses GPT-4o~\citep{hurst2024gpt} to score and create binary preference datasets. However, this approach incurs additional costs and relies on external tools like GPT-4o. Similarly, MMedPO~\citep{zhu2024mmedpo} leverages multi-agent systems to construct preference data, but this requires significant computational resources and memory overhead. In clinical medical settings, there is a pressing need for a cost-efficient method to implement preference optimization, enhancing both the trustworthiness and modality alignment of medical VLMs.

\section{Conclusion}
This paper proposes Hierarchical Self-Contrastive Rewarding (HSCR) to address modality misalignment in medical VLMs. By leveraging visual token dropout, alignment rewards, and dual preference optimization, HSCR improves modality alignment, trustworthiness, and zero-shot performance with minimal training data. Our work paves the way for developing trustworthy medical AI systems.

\section{Limitations}
While our work demonstrates promising results, there are some limitations that warrant further exploration. First, the limited availability of high-quality and diverse medical data continues to constrain the development of medical Vision-Language Models (VLMs). This scarcity impacts the generalizability and robustness of the models, particularly in addressing rare or complex medical scenarios. Second, although our evaluation covers extensive experimental benchmarks, it is primarily conducted in controlled research settings. This leaves room for further validation through integration with clinical workflows and real-world trials to better assess the practicality, reliability, and safety of the proposed methods in healthcare applications.

\section*{Acknowledgements}
This work is supported by the National Natural Science Foundation of China (Grant No. 12326612, 62476241), the Natural Science Foundation of Zhejiang Province, China (Grant No. LZ23F020008), and the Zhejiang University-Angelalign Inc. R\&D Center for Intelligent Healthcare.

\bibliography{custom}

\appendix
\section{Appendix}
\label{sec:appendix}

\subsection{Details of Figure 1}
\label{details:fig1}
In Figure~\ref{fig:intro1}(a), we construct dispreferred responses by applying GPT-4o~\cite{hurst2024gpt} and LLaVA-Med1.5~\citep{li2024llava} separately to paired images and captions from the PMC-15M dataset~\citep{zhang2023large}. Preference optimization is then conducted following prior works~\citep{sun2023aligning,zhou2024aligning,jiang2024modality}. The bar chart on the right illustrates the sampling probability distribution of preference datasets constructed using these two methods. It is evident that preference pairs generated by GPT-4o exhibit significantly lower sampling probabilities compared to those generated by medical VLMs, with most probabilities falling below 0.5. In contrast, the majority of VLM-generated data have sampling probabilities exceeding 0.6. This discrepancy suggests that external preferences derived from more powerful models are inconsistent with the inherent preferences embedded in medical VLMs' own generations, making them suboptimal for preference optimization datasets.

In Figure~\ref{fig:intro1}(b), we manually modify the original image-caption pairs from the PMC-15M dataset to introduce varying levels of quality, thereby creating multi-level dispreferred responses. These dispreferred responses, alongside the original correct captions, are used to construct preference pairs for preference optimization. When evaluated on the PathVQA~\citep{he2020pathvqa} benchmark, the results are unsatisfactory, with no significant correlation observed between model performance and the quality levels of dispreferred responses. This indicates that the model fails to effectively learn distinctions between high-quality preference datasets.
\subsection{Details of Preliminaries}
\label{details:pre}
\noindent \textbf{Reinforcement Learning from Human Feedback (RLHF)}
Reinforcement Learning from Human Feedback (RLHF) is a widely used framework for aligning models with human preferences. It involves training a reward model \( f_{\xi} \) on pairwise preference data~\citep{bai2022training,knox2011augmenting,christiano2017deep}. The reward model is optimized using a cross-entropy loss function:
\begin{equation}
\label{RLHFRM}
    \mathcal{J}_{\text{Reward}} = -\log \left( \sigma \left( f_{\xi}(x, y_w) - f_{\xi}(x, y_l) \right) \right),
\end{equation}
where $\sigma(\cdot)$ is the logistic sigmoid function, and \( f_{\xi}(x, y_w) \) and \( f_{\xi}(x, y_l) \) represent the rewards assigned to the preferred and less preferred outputs, respectively. Once the reward model is trained, the policy \( \pi_{\eta} \) is optimized to maximize the expected reward while maintaining proximity to a reference policy \( \pi_{\text{base}} \). This is achieved through the following objective:
\begin{equation}
\resizebox{0.5\textwidth}{!}{$
    \max_{\pi_{\eta}} \mathbb{E}_{x \sim \mathcal{P}, y \sim \pi_{\eta}(y|x)} \left[ f_{\xi}(x, y) - \gamma D_{\text{KL}}(\pi_{\eta}(y|x) \parallel \pi_{\text{base}}(y|x)) \right],
    $}
\end{equation}
where \( \gamma \) controls the trade-off between reward maximization and regularization, and \( D_{\text{KL}} \) is the Kullback-Leibler divergence~\citep{van2014renyi}, which ensures that policy remains stable and does not deviate excessively from the reference policy~\citep{peng2023stabilizing,zhu2024iterative}.

\noindent \textbf{Direct Preference Optimization (DPO)}
Direct Preference Optimization (DPO) offers a streamlined alternative to RLHF by directly optimizing the policy \( \pi_{\eta} \) using preference data \( \mathcal{P} \), thereby bypassing the need for an explicit reward model~\citep{rafailov2024direct}. DPO establishes a direct relationship between the reward function \( g(x, y) \) and the policy \( \pi_{\eta} \), expressed as:
\begin{equation}
\label{DPOr}
    g(x, y) = \gamma \log \frac{\pi_{\eta}(y|x)}{\pi_{\text{base}}(y|x)} + \gamma \log Z(x),
\end{equation}
where \( Z(x) \) is the partition function that ensures normalization. By substituting this relationship into the reward model loss, DPO formulates the following optimization objective:
\begin{equation}
\resizebox{0.5\textwidth}{!}{$
\begin{aligned}
   \mathcal{J}_{\text{DPO}}(\pi_{\eta}; \pi_{\text{init}}) &= - \mathbb{E}_{(x, y_w, y_l) \sim \mathcal{P}} \left[ \log \sigma \left( \gamma \log \frac{\pi_{\eta}(y_w|x)}{\pi_{\text{init}}(y_w|x)} \right. \right. \\
   &\quad \left. \left. - \gamma \log \frac{\pi_{\eta}(y_l|x)}{\pi_{\text{init}}(y_l|x)} \right) \right], 
\end{aligned}
$}
\end{equation}
where \( y_w \) and \( y_l \) denote the preferred and less preferred outputs, respectively. This approach eliminates the complexity of training a separate reward model, enabling more efficient and precise alignment with human preferences~\citep{dong2024rlhf,pal2024smaug}.

In the context of Vision Language Models (VLMs), DPO typically involves combining the image \( m \) and textual query \( t \) into a unified input \( x \)~\citep{jiang2024modality,yu2024rlaif,sun2023aligning,zhu2024mmedpo}. The preferred responses \( y_w \) are selected for their accuracy and relevance, while the less preferred responses \( y_l \) often contain errors or irrelevant information.

\begin{table*}[!ht]
\centering
\resizebox{1\textwidth}{!}{%
\begin{tabular}{cc|cc|cc|cc}
\toprule
\textbf{Mask Ratio} & & \multicolumn{2}{c|}{\cellcolor[HTML]{E8F5E9}\textbf{RAD-VQA}} & \multicolumn{2}{c|}{\cellcolor[HTML]{FCE4EC}\textbf{SLAKE}} & \multicolumn{2}{c}{\cellcolor[HTML]{FFF9E3}\textbf{PathVQA}} \\
 & & \cellcolor[HTML]{E8F5E9}\textbf{Open} & \cellcolor[HTML]{E8F5E9}\textbf{Closed} & \cellcolor[HTML]{FCE4EC}\textbf{Open} & \cellcolor[HTML]{FCE4EC}\textbf{Closed} & \cellcolor[HTML]{FFF9E3}\textbf{Open} & \cellcolor[HTML]{FFF9E3}\textbf{Closed} \\
\midrule
Baseline & & \cellcolor[HTML]{E8F5E9}32.31 & \cellcolor[HTML]{E8F5E9}56.62 & \cellcolor[HTML]{FCE4EC}42.45 & \cellcolor[HTML]{FCE4EC}56.49 & \cellcolor[HTML]{FFF9E3}10.01 & \cellcolor[HTML]{FFF9E3}59.75 \\
0.3 & & \cellcolor[HTML]{E8F5E9}33.21 (+0.90) & \cellcolor[HTML]{E8F5E9}57.65 (+1.03) & \cellcolor[HTML]{FCE4EC}43.33 (+0.88) & \cellcolor[HTML]{FCE4EC}59.94 (+3.45) & \cellcolor[HTML]{FFF9E3}10.34 (+0.33) & \cellcolor[HTML]{FFF9E3}61.04 (+1.29) \\
0.5 & & \cellcolor[HTML]{E8F5E9}33.35 (+1.04) & \cellcolor[HTML]{E8F5E9}58.76 (+2.14) & \cellcolor[HTML]{FCE4EC}44.47 (+2.02) & \cellcolor[HTML]{FCE4EC}61.83 (+5.34) & \cellcolor[HTML]{FFF9E3}11.96 (+1.95) & \cellcolor[HTML]{FFF9E3}62.82 (+3.07) \\
0.9 & & \cellcolor[HTML]{E8F5E9}34.21 (+1.90) & \cellcolor[HTML]{E8F5E9}59.65 (+3.03) & \cellcolor[HTML]{FCE4EC}44.35 (+1.90) & \cellcolor[HTML]{FCE4EC}62.77 (+6.28) & \cellcolor[HTML]{FFF9E3}12.15 (+2.14) & \cellcolor[HTML]{FFF9E3}62.67 (+2.92) \\
0.7 & & \cellcolor[HTML]{E8F5E9}35.92 (+3.61) & \cellcolor[HTML]{E8F5E9}60.13 (+3.51) & \cellcolor[HTML]{FCE4EC}45.32 (+2.87) & \cellcolor[HTML]{FCE4EC}63.46 (+6.97) & \cellcolor[HTML]{FFF9E3}12.36 (+2.35) & \cellcolor[HTML]{FFF9E3}64.17 (+4.42) \\
\bottomrule
\end{tabular}%
}
\caption{Performance comparison with different mask ratios.}
\label{tab:mask_ratio}
\end{table*}

\subsection{Details of Evaluation Datasets}
\label{ref:evaldata}
\noindent\textbf{VQA-RAD}~\citep{lau2018dataset} is a medical visual question answering dataset consisting of 3,515 question-answer pairs. The questions span 11 distinct categories and include both closed-ended (e.g., yes/no) and open-ended (e.g., descriptive) types, providing a comprehensive evaluation of model capabilities in medical contexts.

\noindent\textbf{SLAKE}~\citep{liu2021slake} is a multimodal dataset designed for medical VQA, featuring over 7,000 question-answer pairs. It includes detailed annotations such as semantic segmentation masks and object detection bounding boxes, enabling fine-grained visual understanding. The dataset covers diverse anatomical regions, including the brain, neck, chest, abdomen, and pelvic cavity, and supports both English and Chinese. For consistency, our experiments focus exclusively on the English subset.

\noindent\textbf{PathVQA}~\citep{he2020pathvqa} is a pathology-focused dataset containing 4,998 images paired with 32,799 question-answer pairs. The questions address various attributes, including spatial location, morphological features (e.g., shape, color), and pathological characteristics. These questions are categorized into open-ended and closed-ended types, offering a robust benchmark for evaluating model performance in pathology-related reasoning tasks.

% \subsection{Details of Training}
% \label{ref:expe_detail}
% We sample 2,000 entries from the dataset used in instruction-tuning stage to construct preference datasets as described in Methods~\ref{methods1}. For this process, we set the parameters as follows: \( j = 3 \), \( \beta = 0.9 \), \( n = 10 \), and \( \gamma = 0.1 \). Multi-level preference optimization is performed using the constructed implicit and explicit preference datasets, with LoRA~\citep{hu2021lora} applied at a rank of 16. Training is conducted for 2 epochs with a learning rate of 5e-7 and without weight decay. The training hyperparameters are in Table~\ref{tab:setting}.

\begin{table*}[ht!]
\centering
\begin{tabular}{l|c|c|c}
\toprule
\textbf{Config} & \multicolumn{3}{c}{HSCR Training Config} \\
\midrule
Deepspeed & \multicolumn{3}{c}{Zero2} \\
Image encoder & \multicolumn{3}{c}{ CLIP-ViT-L/14@336px } \\
Feature select layer & \multicolumn{3}{c}{-2} \\
Image projector & \multicolumn{3}{c}{2 Linear layers with GeLU} \\
Epoch &\multicolumn{3}{c}{2} \\
Learning rate &\multicolumn{3}{c}{5e-7} \\
Learning rate schedule & \multicolumn{3}{c}{Cosine} \\
Weight decay & \multicolumn{3}{c}{0.0} \\
Text max length & \multicolumn{3}{c}{4096} \\
Batch size per GPU & \multicolumn{3}{c}{2} \\
GPU & \multicolumn{3}{c}{8 × 3090-24G} \\
Precision & \multicolumn{3}{c}{Bf16} \\
\bottomrule
\end{tabular}
\caption{Our experimental hyperparameters.}
\label{tab:setting}
\end{table*}

\subsection{Comparison with LiPO}
\label{ref:mask}
While both our method and LiPO~\citep{liu2024lipo} leverage list preference datasets for optimization, they differ significantly in scope, data generation, and training requirements. LiPO primarily focuses on RLHF for text-only models and tasks, whereas our approach, HSCR, is specifically designed for Vision-Language Models (VLMs) in medical multimodal tasks, aiming to enhance trustworthiness and modality alignment. In terms of preference data generation, LiPO employs prompt-based methods to create sentence-level dispreferred responses that differ substantially from the preferred ones, enabling easier distinction early in optimization. In contrast, HSCR generates token-level preference data through visual token dropout and contrastive decoding, producing dispreferred responses that remain largely similar to the preferred ones while introducing critical divergences that reflect misalignment, thereby enabling finer-grained optimization. Regarding training requirements, LiPO relies on a reward model (T5-XXL, 11B parameters) trained with human-annotated datasets for ranking preference lists, which introduces significant computational overhead and depends heavily on large-scale human annotations—a resource particularly scarce in the medical domain. In contrast, HSCR eliminates the need for external MLLMs or human annotations by leveraging the model’s internal behavior to generate and rank preference data, achieving superior performance with minimal cost and requiring only 2,000 training samples.

\subsection{Details of Mask Strategies}
\label{ref:lipo}
\noindent\textbf{Pixel-level Masking:} In this approach, we directly crop 70\% of the content from the original image, leaving only 30\% of the pixels as input to the CLIP visual encoder. This technique is commonly employed in many previous VLMs.

\noindent\textbf{Patch-level Masking:} Given that the patch size of the CLIP visual encoder in LLaVA-Med is 14, we divide the original image into patches of size 14 and randomly discard 70\% of them. As a result, only the remaining 30\% of patches are fed into the Vision Transformer (ViT).

\noindent\textbf{Latent-space Masking:} In this strategy, we modify the attention mask within the CLIP visual encoder of LLaVA-Med by randomly setting 70\% of the mask values to -inf, simulating a latent-space level mask.

\noindent\textbf{Visual Token Dropout:} This technique disrupts the latent space after the projector in the VLM. Specifically, the image is first encoded by the ViT, then transformed into visual tokens by an MLP projector, and finally concatenated with text tokens before being input into the LLM backbone. We drop 70\% of the visual tokens to induce misalignment responses in the VLMs as they are restricted in their access to visual tokens.

\begin{table*}[htbp!]
\centering
\label{tab:performance}
\begin{tabular}{|l|cccc|}
\hline
\textbf{Method}               & \textbf{CHAIR$_s$ $\downarrow$} & \textbf{Object Coverage $\uparrow$} & \textbf{Hallucination Rate $\downarrow$} & \textbf{Cognition $\downarrow$} \\ \hline
LLaVA-v1.5-7B        & 7.8                    & 51.0                       & 36.4                            & 4.2                    \\
LLaVA-v1.5-7B + DPO  & 6.5                    & 49.1                       & 34.5                            & 3.4                    \\
LLaVA-v1.5-7B + HSCR & \textbf{4.4}                & \textbf{55.5}                   & \textbf{25.1}                        & \textbf{2.9}                \\ \hline
\end{tabular}
\caption{Performance comparison on general multimodal benchmark.}
\label{tab:app_general}
\end{table*}

\subsection{Implementation Details of Figure 7}
\label{ref:fig7}
We evaluated LLaVA-v1.5-7B on the widely adopted AMBER benchmark~\citep{wang2023llm}, maintaining consistent experimental parameters with our main experiments. The preference datasets were constructed using HSCR based on the RLHF-V dataset~\citep{yu2024rlhf}. The AMBER benchmark comprehensively assesses VLM trustworthiness through four key metrics:

\begin{itemize}
\item \textbf{CHAIR}: Quantifies object hallucination frequency in generated captions (lower values indicate better performance).
\item \textbf{Object Coverage}: Measures the proportion of image objects accurately described in captions (higher values indicate better performance).
\item \textbf{Hallucination Rate}: Evaluates the frequency of hallucinated objects in generated descriptions (lower values indicate better performance).
\item \textbf{Cognition}: Measures the degree to which hallucinations in VLMs align with human cognitive patterns. This metric evaluates whether the model's hallucination tendencies resemble those observed in human cognition processes (lower values indicate better performance).
\end{itemize}
\begin{figure*}[!t]
    \centering
    \includegraphics[width=1\linewidth]{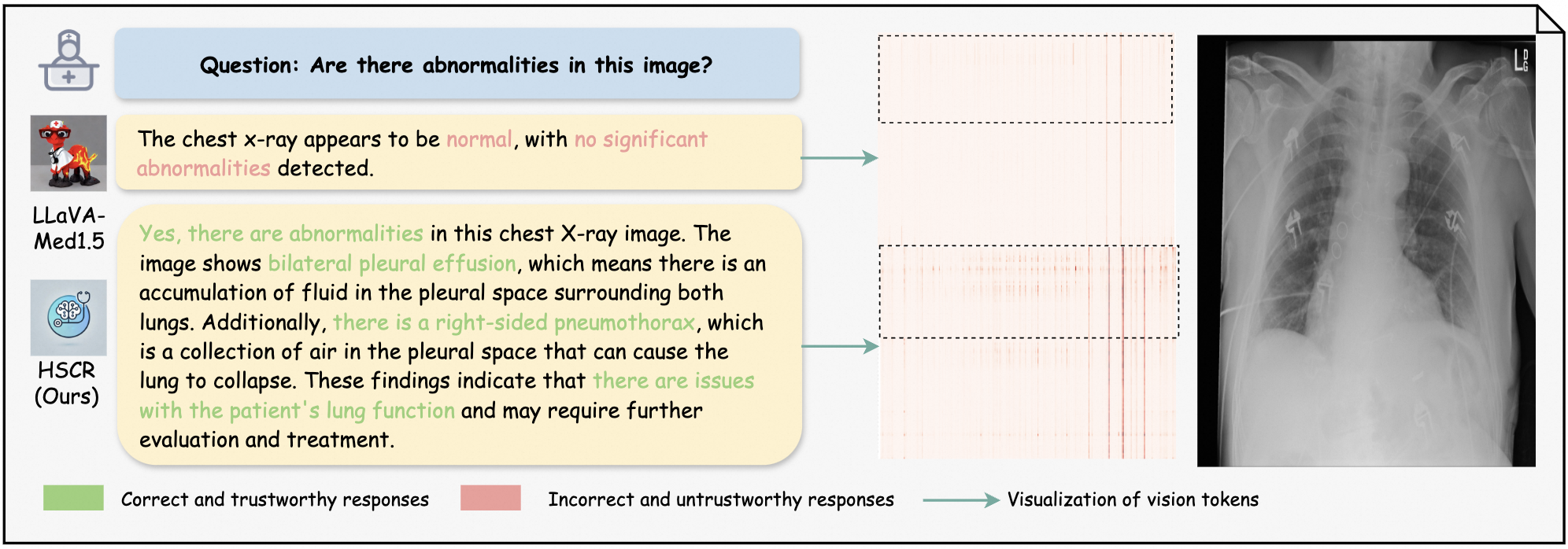}
    \includegraphics[width=1\linewidth]{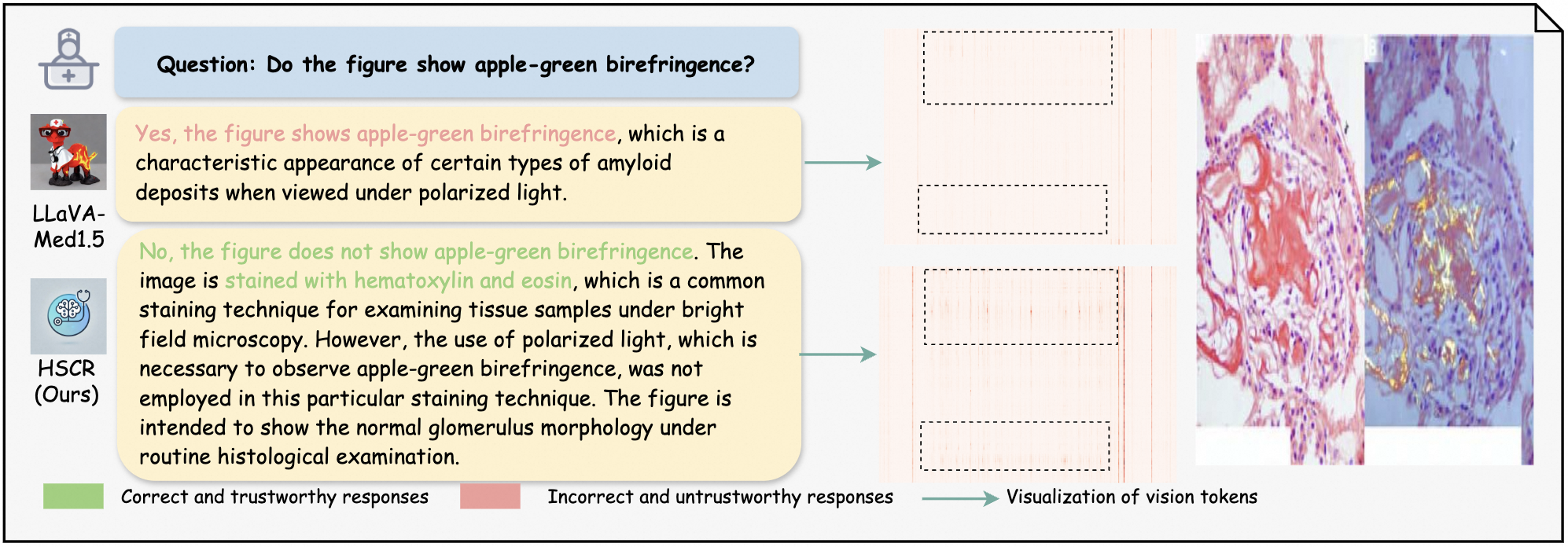}
\caption{More cases on zero-shot comparison of LLaVA-Med1.5~\citep{li2024llava} and HSCR.}
   \label{fig:more_case}
\end{figure*}

\end{document}